\documentclass[runningheads]{llncs}
\usepackage{graphicx}
\usepackage{comment}
\usepackage{amsmath,amssymb} 
\usepackage{color}
\usepackage{times}
\usepackage{epsfig}
\usepackage{graphicx}
\usepackage{amsmath}
\usepackage{amssymb}
\usepackage{algorithm}
\usepackage{algpseudocode}
\usepackage{xcolor}
\usepackage{multirow}
\usepackage{caption}
\usepackage{makecell}

\newcommand{\eg}{{\textit{e.g.}}}
\newcommand{\etal}{{\textit{et al.}}}

\newcommand{\ie}{{\textit{i.e.}}}

\begin{document}
\pagestyle{headings}
\mainmatter
\def\ECCVSubNumber{5204}  

\title{VCNet: A Robust Approach to Blind Image Inpainting}

\titlerunning{VCNet: A Robust Approach to Blind Image Inpainting}
%
\author{Yi Wang\inst{1} \and
Ying-Cong Chen\inst{1} \and Xin Tao \inst{2} and
Jiaya Jia\inst{1,3}}
\authorrunning{Y. Wang, Y. C. Chen, X. Tao, J. Jia}
%
\institute{$^{1}$The Chinese University of Hong Kong \quad $^{2}$YouTu Lab, Tencent \quad $^{3}$SmartMore\\
\email{\{yiwang, ycchen, leojia\}@cse.cuhk.edu.hk} \quad \email{xintao@tencent.com}}
\maketitle

\begin{center}
	\centering
	\includegraphics[width=1\linewidth]{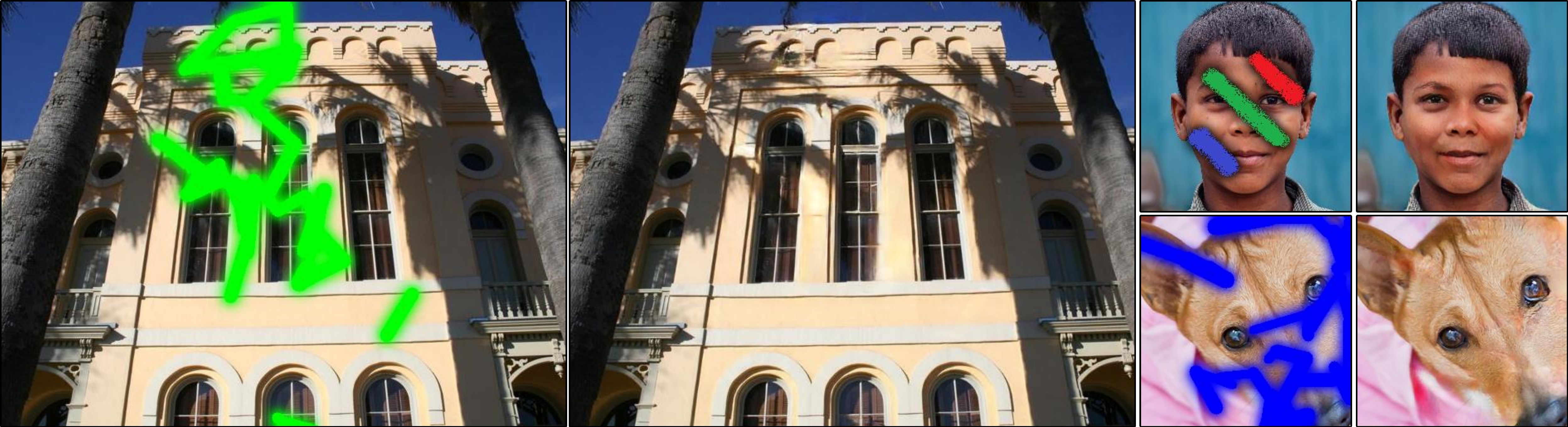}
	\small
	\captionof{figure}{Our blind inpainting method on scenes (\textbf{left}, from Places2 \cite{zhou2017places}), face (\textbf{top right}, from FFHQ \cite{karras2018style}), and animal (\textbf{bottom right}, from ImageNet \cite{deng2009imagenet}). \textit{No} masks are provided during inference, and these mask filling patterns are not included in our training.}
	\label{fig:teaser}
\end{center}
	\vspace{-0.2in}

\begin{abstract}
Blind inpainting is a task to automatically complete visual contents without specifying masks for missing areas in an image. Previous works assume missing region patterns are known, limiting its application scope. In this paper, we relax the assumption by defining a new blind inpainting setting, making training a blind inpainting neural system robust against various unknown missing region patterns. Specifically, we propose a two-stage visual consistency network (VCN), meant to estimate where to fill (via masks) and generate what to fill. In this procedure, the unavoidable potential mask prediction errors lead to severe artifacts in the subsequent repairing. To address it, our VCN predicts semantically inconsistent regions first, making mask prediction more tractable. Then it repairs these estimated missing regions using a new spatial normalization, enabling VCN to be robust to the mask prediction errors. In this way, semantically convincing and visually compelling content is thus generated. Extensive experiments are conducted, showing our method is effective and robust in blind image inpainting. And our VCN allows for a wide spectrum of applications.
\end{abstract}

\section{Introduction}

Image inpainting aims to repair missing regions of an image based on its context. Generally, it takes a corrupted image as well as a mask that indicates missing pixels as input, and restore it based on the semantics and textures of uncorrupted regions. It serves applications of object removal, image restoration, etc. We note the requirement of having accurate masks makes it difficult to be practical in several scenarios where masks are not available. Users need to carefully locate corrupted regions manually, where inaccurate masks may lead to inferior results. We in this paper analyze blind inpainting that automatically finds pixels to complete, and propose a suitable solution based on image context understanding.

Existing work \cite{cai2017blind,liu2017blind} on blind inpainting assumes that the missing areas are filled with constant values or Gaussian noise. Thus the corrupted areas can be identified easily and almost perfectly based on noise patterns. This oversimplified assumption could be problematic when corrupted areas are with unknown content. To improve the applicability, we relax the assumption and propose the {\it versatile blind inpainting} task. We solve it by taking deeper semantics of the input image into overall consideration and detecting more semantically meaningful \textit{inconsistency} based on the structural context in contrast to previous blind inpainting. 

Note that blind inpainting without assuming the damage patterns is highly ill-posed. This is because the unknown degraded regions need to be located based on their differences from the intact ones instead of their known characteristics, and the uncertainties in this prediction make the further inpainting challenging. We address it in two aspects, i.e., a new data generation approach and a novel network architecture. 

For training data collection, if we only take common black or noise pixels in damaged areas as input, the network may detect these patterns as features instead of utilizing the contextual semantics as we need. In this scenario, the damage for training should be diverse and complicated enough so that the high-level of structure inconsistency instead of the pattern in damage can be extracted. Our first contribution, therefore, is the new strategy to generate highly diverse training data where natural images are adopted as the filling content with random strokes. 

For model design, our framework consists of two stages as mask prediction and robust inpainting. A discriminative model is used to conduct binary pixel-wise classification to predict inconsistent areas. With the mask estimated, we use it to guide the inpainting process. Though this framework is intuitive, its specific designs to address the biggest issue in this framework are non-trivial: \textit{how to neutralize the generation degradation brought by inevitable mask estimation errors} in the first stage. To cope with this challenge, we propose a probabilistic context normalization (PCN) to spatially transfers contextual information in different neural layers, enhancing information aggregation of the inpainting network based on the mask prediction probabilities. We experimentally validate it outperforms other existing approaches exploiting masks, e.g., concatenating mask with the input image and using convolution variants (like Partial Convolution \cite{liu2018image} or Gated Convolution \cite{yu2019free}) to employ masks, in the generation evaluations.

Though trained without seeing any graffiti or trivial noise patterns (\eg\ constant color or Gaussian noise), our model can automatically remove them without manually annotated marks, even for complex damages introduced by real images. This is validated in several benchmarks like FFHQ \cite{karras2018style}, ImageNet \cite{deng2009imagenet}, and Places2 \cite{zhou2017places}. Besides, we find our predicted mask satisfyingly focuses on visual inconsistency in images as expected instead of inherent damage patterns when these two stages are jointly trained in an adversarial manner. This further improves robustness for this very challenging task, and leads to the application of exemplar-guided face-swap (Sec. \ref{sec_face_swap}).  Also, such blind inpainting ability can be transferred to other removal tasks such as severe raindrop removal as exemplified in Sec. \ref{sec_exp}. Many applications are enabled.

Our contribution is twofold. First, we propose the first relativistic generalized blind inpainting system. It is robust against various unseen degradation patterns and mask prediction errors. We jointly model mask estimation and inpainting procedure, and address error propagation from the computed masks to the subsequent inpainting via a new spatial normalization. Second, effective tailored training data synthesis for this new task is presented with comprehensive analysis. It makes our blind inpainting system robust to visual inconsistency, which is beneficial for various inpainting tasks. Our system finds many practical applications.

\section{Related Work}
\vspace{-0.1in}
\paragraph{\textbf{Blind Image Inpainting}}
Conventional inpainting methods employ external or internal image local information to fill the missing regions \cite{barnes2009patchmatch,criminisi2004region,darabi2012image,jia2003image,kopf2012quality,levin2003learning,sun2005image}. For the blind image setting, existing research \cite{cai2017blind,liu2017blind} assumes contamination with simple data distributions, \eg\ text-shaped or thin stroke masks filled with constant values. This setting makes even a simple model applicable by only considering local information, without the need for a high-level understanding of the input.

\vspace{-0.15in}
\paragraph{\textbf{Generative Image Inpainting}} 
Recent advance \cite{arjovsky2017wasserstein,miyato2018spectral,zhang2018self,wang2018high} in the conditional generative models makes it possible to fill large missing areas in images \cite{pathak2016context,iizuka2017globally,li2017generative,yang2017high,yu2018generative,wang2018inpainting,xiong2019foreground,xie2019image,zheng2019pluralistic,sagong2019pepsi}. Pathak \etal\ \cite{pathak2016context} learned an inpainting encoder-decoder network using both reconstruction and adversarial losses.
Iizuka \etal\ \cite{iizuka2017globally} proposed the global and local discriminators for the adversarial training scheme. 
To obtain more vivid texture, coarse-to-fine \cite{yang2017high,yu2018generative,xiong2019foreground} or multi-branch \cite{wang2018inpainting} network architecture, and non-local patch-match-like losses \cite{yang2017high,wang2018inpainting} or network layer \cite{yu2018generative} were introduced. 

Specifically, Yang \etal\ \cite{yang2017high} applied style transfer in an MRF manner to post-process the output of the inpainting network, creating crisp textures at the cost of heavy iterative optimization during testing. Further, Yu \etal\ \cite{yu2018generative} conducted the neural patch copy-paste operation with full convolutions, enabling texture generation in one forward pass. Instead of forcing copy-paste in the testing phase, Wang \etal\ \cite{wang2018inpainting} gave MRF-based non-local loss to encourage the network to model it implicitly. To better handle the generation of the missing regions, various types of intermediate representations (\eg\ edges \cite{nazeri2019edgeconnect} and foreground segmentation \cite{xiong2019foreground}) are exploited to guide the final fine detail generation in a two-stage framework. Meanwhile, some researches focus on generating pluralistic results \cite{zheng2019pluralistic} or improving generation efficiency \cite{sagong2019pepsi}.

Also, research exists to study convolution variants \cite{ren2015shepard,uhrig2017sparsity,liu2018image,yu2019free}. They exploit the mask more explicitly than simple concatenation with the input. Generally, the crafted networks learn upon known pixels indicated by the mask.

\vspace{-0.15in}
\paragraph{\textbf{Other Removal Tasks}} A variety of removal tasks are related to blind inpainting, \eg\, raindrop removal \cite{qian2018attentive}. Their assumptions are similar regarding the condition that some pixels are clean or useful. The difference is on feature statistics of noisy areas subjects to some strong priors.

\begin{figure*}[t]
	\centering
	\includegraphics[width=0.9\linewidth]{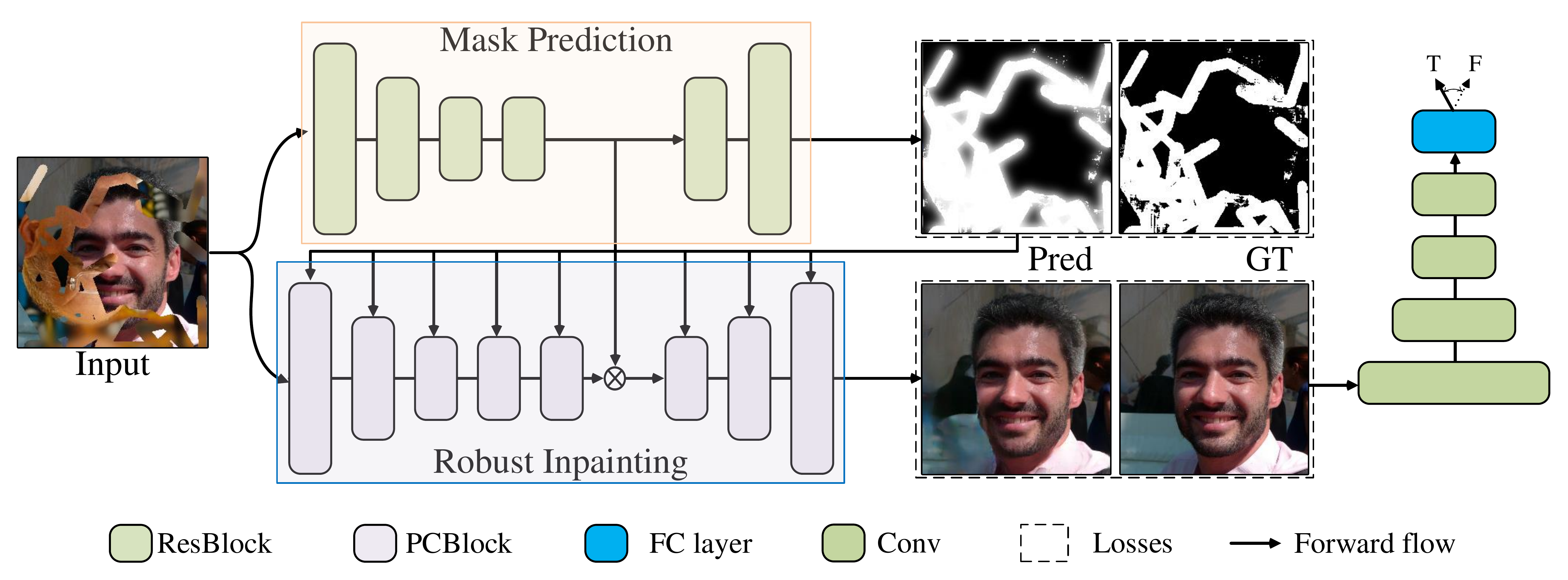}
	\vspace{-0.1in}
	\caption{Our framework. It consists of sequentially connected mask prediction and robust inpainting networks, trained in an adversarial fashion.}
	\label{fig:blind-inpaintingframework}
\end{figure*}
	\vspace{-0.1in}

\section{Robust Blind Inpainting} \label{method}

For this task, the input is only a degraded image $\mathbf{I} \in \mathbb{R}^{h \times w \times c}$ (contaminated by unknown visual signals), and the output is expected to be a plausible image $\mathbf{\hat{O}} \in \mathbb{R}^{h \times w \times c}$, approaching ground truth $\mathbf{O} \in \mathbb{R}^{h \times w \times c}$ of $\mathbf{I}$.

The degraded image $\mathbf{I}$ in the blind inpainting setting is formulated as
\begin{equation} \label{degrade_model}
\mathbf{I} = \mathbf{O} \odot (\mathbf{1}-\mathbf{M}) + \mathbf{N} \odot \mathbf{M},
\end{equation}
where $\mathbf{M} \in \mathbb{R}^{h \times w \times 1}$ is a binary region mask (with value 0 for known pixels and 1 otherwise), and $\mathbf{N} \in \mathbb{R}^{h \times w \times c}$ is an noisy visual signal. $\odot$ is the Hadamard product operator. Given $\mathbf{I}$, we need to predict $\mathbf{\hat{O}}$ (an estimate of $\mathbf{O}$) with latent variables $\mathbf{M}$ and $\mathbf{N}$. Also, Eq. \eqref{degrade_model} is the means we use to produce training tuples $<\mathbf{I}_i, \mathbf{O}_i, \mathbf{M}_i, \mathbf{N}_i>_{|i=1,...,m}$.

\subsection{Training Data Generation} \label{sec_data_gen}

How to define the possible image contaminations ($\mathbf{N}$ indicates what and $\mathbf{M}$ indicates where in Eq. \eqref{degrade_model}) is the essential prerequisite for whether a neural system could be robust to a variety of possible image contaminations. Setting $\mathbf{N}$ as a constant value or certain kind of noise makes it and $\mathbf{M}$ easy to be distinguished by a deep neural net or even a simple linear classifier from a natural image patch. This prevents the model to predict inconsistent regions based on the semantic context, as drawing prediction with the statistics of a local patch should be much easier. It converts the originial blind inpainting problem into a vanilla inpainting one with a nearly perfect prediction of $\mathbf{M}$. It becomes solvable with the existing techniques, but its assumption generally does not hold in the real-world scenarios, \eg, graffiti removal shown in Fig. \ref{fig:teaser}.

 In this regard, the key for defining $\mathbf{N}$ is to make it indistinguishable as much as possible with $\mathbf{I}$ on image pattern, so that the model cannot decide if a local patch is corrupted without seeing the semantic context of the image. Then a neural system trained with such data has the potential to work on the unknown contaminations.

In this paper, we use real-world image patches to form $\mathbf{N}$. This ensures that local patches between $\mathbf{N}$ and $\mathbf{I}$ are indistinguishable, enforcing the model to draw an inference based on contextual information, which eventually improves the generalization ability for real-world data. Further, we alleviate any priors introduced by $\mathbf{M}$ in training via employing free-form strokes \cite{yu2019free}. Existing blind or non-blind inpainting methods often generate the arbitrary size of a rectangle or text-shaped masks. However, this is not suitable for our task, because it may encourage the model to locate the corrupted part based on the rectangle shape. Free-form masks can largely diversify the shape of masks, making the model harder to infer corrupted regions with shape information.

Also, it is worth noting that direct blending image $\mathbf{O}$ and $\mathbf{N}$ using Eq. \eqref{degrade_model} would lead to noticeable edges, which are strong indicators to distinguish noisy areas. This will inevitably sacrifice the semantic understanding capability of the used model. Thus, we dilate the $\mathbf{M}$ into $\mathbf{\tilde{M}}$ by the iterative Gaussian smoothing in \cite{wang2018inpainting} and employ alpha blending in the contact regions between $\mathbf{O}$ and $\mathbf{N}$.

\subsection{Our Method}

We propose an end-to-end framework, named Visual Consistent Network (VCN) (Fig. \ref{fig:blind-inpaintingframework}). VCN has two submodules, \ie\, Mask Prediction Network (MPN) and Robust Inpainting Network (RIN). MPN is to predict potential visually inconsistent areas of a given image, while RIN is to inpaint inconsistent parts based on the predicted mask and the context in the original image. Note that these two submodules are correlated. MPN provides an inconsistency mask $\mathbf{\hat{M}} \in \mathbb{R}^{h \times w \times 1}$, where $\mathbf{\hat{M}}_p \in [0, 1]$, helping RIN locate inconsistent regions. On the other hand, by leveraging local and global semantic context, RIN largely regularizes MPN, enforcing it to focus on these regions instead of simply fitting our generated data.

Our proposed VCN is a robust solution to blind image inpainting in the given relativistic generalized setting. Its robustness is shown in two aspects. MPN of VCN can predict the regions to be repaired with decent performance even the contamination patterns are novel to the trained model. More importantly, RIN of VCN synthesizes plausible and convincing visual contents for the predicted missing regions, robust against the mask prediction errors. Their specific designs are detailed below.

\vspace{-0.1in}
\paragraph{\textbf{Mask Prediction Network (MPN)}} MPN aims to learn a mapping $F$ where $F(\mathbf{I}) \rightarrow \mathbf{M}$. MPN is with an encoder-decoder structure using residual blocks \cite{he2016deep}, and takes binary cross entropy loss between $\mathbf{\hat{M}}$ and $\mathbf{M}$ as the optimization goal. To stabilize its learning, a self-adaptive loss is introduced to balance positive- and negative-sample classification, because clear pixels outnumber the damages ones ($|\{p|\mathbf{M}_p=1\}| = \rho |\{p|\mathbf{M}_p=0\}|$ where $\rho=0.56 \pm 0.17$). This self-adaptive loss is expressed as
\begin{equation} \label{weighted_bce}
\mathcal{L}_m(\mathbf{\hat{M}}, \mathbf{M}) = -\tau \sum_{p}\mathbf{M}_p \cdot \log(\mathbf{\hat{M}}_p)  - (1-\tau) \sum_{q}(\mathbf{1}-\mathbf{M}_q) \cdot \log(\mathbf{1}-\mathbf{\hat{M}}_q),
\end{equation}
where $p \in \{p | \mathbf{M}_p = 1 \}$, $q \in \{q | \mathbf{M}_q = 0 \}$, and $\tau = |\{p|\mathbf{M}_p=0\}| / (h \times w)$.

\begin{figure}[t]
	\begin{center}
		\centering
		\begin{tabular}{ccccc}
			\multicolumn{5}{c}{\includegraphics[width=0.98\linewidth]{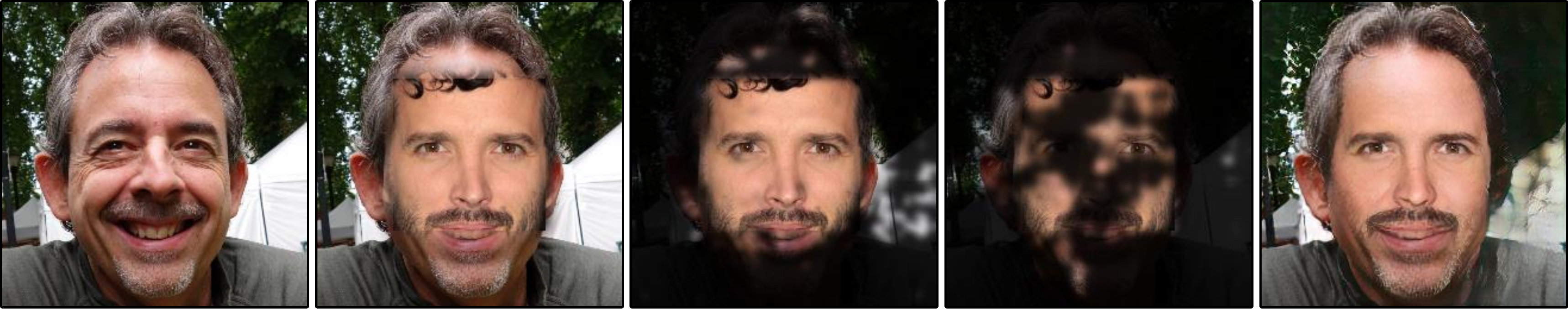}} \\
			\hspace{0.10\columnwidth}(a) & \hspace{0.15\columnwidth}(b) & \hspace{0.16\columnwidth}(c) & \hspace{0.16\columnwidth}(d) & \hspace{0.08\columnwidth}(e) 
		\end{tabular}
	\end{center}
	\vspace{-0.15in}
	\caption{Visualization of learned masks with different training strategies. (a) The input ground truth. (b) Input face image whose central part is replaced by another face with the rectangle mask. (c) Estimated mask with training MPN alone. (d) Estimated mask with joint training with inpainting network. (e) Output image of VCN. The used MPN is trained with free-form stroke masks \cite{yu2019free}.} 
	\label{fig_masklearning}
\end{figure}

Note $\mathbf{\hat{M}}$ is an estimated soft mask where $0 \le \mathbf{\hat{M}}_p \le 1$ for $\forall p$, although we employ a binary version for $\mathbf{M}$ in Eq. \eqref{degrade_model}. It means the damaged pixels are not totally abandoned in the following inpainting process. The softness of $\mathbf{\hat{M}}$ enables the differentiability of the whole network. Additionally, it lessens error accumulation caused by pixel misclassification, since pixels whose status (damaged or not) MPN are uncertain about are still utilized in the later process. 

Note that the objective of MPN is to detect all corrupted regions. Thus it tends to predict large corrupted regions for an input corrupted image, which is shown in Fig. \ref{fig_masklearning}(c). As a result, it makes the subsequent inpainting task too difficult to achieve. 
To make the task more tractable, we instead propose to detect the \textit{inconsistency} region of the image, as shown in Fig. \ref{fig_masklearning}(d), which is much smaller. If these regions are correctly detected, other corrupted regions can be naturally blended to the image, leading to realistic results. In the following, we show that by jointly learning MPN with RIN, the MPN eventually locates \textit{inconsistency} regions instead of all corrupted ones.

\begin{figure}[!t]
	\begin{center}
		\centering
		\begin{tabular}{cc}
			\makecell*[c]{\includegraphics[width=0.48\linewidth]{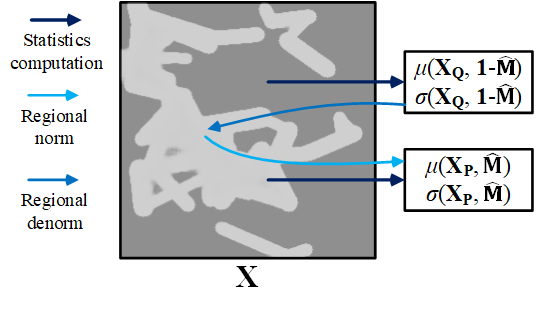}}
			 &
			 \makecell*[c]{\includegraphics[width=0.48\linewidth]{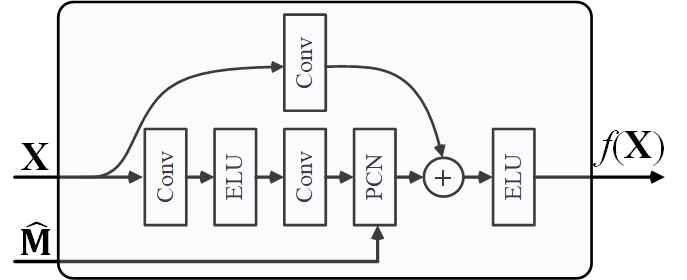}}\\
			\hspace{0.00\columnwidth}(a) & \hspace{0.00\columnwidth}(b) 
		\end{tabular}
	\end{center}
	\vspace{-0.15in}
	\caption{The designs of (a) operator $\mathcal{T}(\cdot)$ in probabilistic context normalization (PCN, defined in Eq. \eqref{eq:cn}) and (b) probabilistic contextual block (PCB). $\mathbf{X}$ and $\mathbf{\hat{M}}$ denote the input feature map and the predicted mask, respectively. $\mathbf{X}_{\mathbf{P}} = \mathbf{X} \odot \mathbf{\hat{M}}$ and $\mathbf{X}_{\mathbf{Q}} = \mathbf{X} \odot (\mathbf{I}-\mathbf{\hat{M}})$.}
	\label{fig:blind-unitv2}
\end{figure}

\vspace{-0.1in}
\paragraph{\textbf{Robust Inpainting Network (RIN)}} With the $\mathbf{\hat{M}}$ located by MPN, RIN corrects them and produces a realistic result $\mathbf{O}$ -- that is, RIN learns a mapping $G$ that $G(\mathbf{I} | \mathbf{\hat{M}}) \to \mathbf{O}$. Also, RIN is structured in an encoder-decoder fashion with probabilistic contextual blocks (PCB). PCB is a residual block variant armed with a new normalization (Fig. \ref{fig:blind-unitv2}), incorporating spatial information with the predicted mask. 

With the predicted mask $\mathbf{\hat{M}}$, repairing corrupted regions requires knowledge inference from contextual information, and being skeptical to the mask for error propagation from the previous stage. A naive solution is to concatenate the mask with the image and feed them to a network. However, this way captures context semantics only in the deeper layers, and does not consider the mask prediction error explicitly. To improve contextual information accumulation and minimize mask error propagation, it would be better if the transfer is done in all building blocks, driven by the estimated mask confidence. Therefore, we propose a probabilistic context normalization (PCN, Fig. \ref{fig:blind-unitv2}) that transfers contextual information in different layers, enhancing information aggregation of the robust inpainting network. 

Our PCN module is composed of the context feature transfer term and feature preserving term. The former transfers mean and variance from known features to unknown areas, both indicated by the estimated soft mask $\mathbf{\hat{M}}$ ($\mathbf{H}$ defined below is its downsampled version). It is a learnable convex combination of feature statistics from the predicted known areas and unknowns ones.
Feature preserving term keeps the features in the known areas (of high confidence) intact.  The formulation of PCN is given as:
\begin{equation} \label{eq:cn} 
\text{PCN}(\mathbf{X}, \mathbf{H})  \! = \! \underbrace{[\beta \! \cdot \! \mathcal{T}(\mathbf{X},\mathbf{H}) \! + \! (1 \!- \! \beta)\mathbf{X} \! \odot \! \mathbf{H}] \! \odot \! \mathbf{H}}_{\text{Context feature transfer}} \! + \! \underbrace{\mathbf{X} \! \odot \! \mathbf{\bar{H}}}_{\text{Feature preserving}},
\end{equation}
and the operator $\mathcal{T}(\cdot)$ is to conduct instance internal statistics transfer as
\begin{equation} \label{eq:op} 
\mathcal{T}(\mathbf{X}, \mathbf{H}) = \frac{\mathbf{X}_{\mathbf{P}}-\mu(\mathbf{X}_{\mathbf{P}},\mathbf{H})}{\sigma(\mathbf{X}_{\mathbf{P}},\mathbf{H})} \cdot \sigma(\mathbf{X}_{\mathbf{Q}},\mathbf{\bar{H}}) + \mu(\mathbf{X}_{\mathbf{Q}},\mathbf{\bar{H}}),
\end{equation}
where $\mathbf{X}$ is the input feature map of PCN, and $\mathbf{H}$ is nearest-neighbor downsampled from $\mathbf{\hat{M}}$, which shares the same height and width with $\mathbf{X}$. $\mathbf{\bar{H}}=\mathbf{1}-\mathbf{H}$ indicates the regions that MPN considers clean. $\mathbf{X}_{\mathbf{P}} = \mathbf{X} \odot \mathbf{H}$ and $\mathbf{X}_{\mathbf{Q}} = \mathbf{X} \odot \mathbf{\bar{H}}$. $\beta$ is a learnable channel-wise vector ($\beta \in \mathcal{R}^{1 \times 1 \times c}$ and $\beta \in [0, 1]$) computed from $\mathbf{X}$ by a squeeze-and-excitation module \cite{hu2018squeeze} as
\begin{equation}
\beta = f(\bar{x}), \quad \text{and} \quad \bar{x}_k=\frac{1}{h' \times w'}\sum_{i=1}^{h'}\sum_{j=1}^{w'}\mathbf{X}_{i,j,k},
\end{equation}
where $\bar{x}\in \mathcal{R}^{1 \times 1 \times c}$ is also a channel-wise vector computed by average pooling $\mathbf{X}$, and $f(\cdot)$ is the excitation function composed by two fully-connected layers with activation functions (ReLU and Sigmoid, respectively).

 $\mu(\cdot, \cdot)$ and $\sigma(\cdot, \cdot)$ in Eq. \eqref{eq:op} compute the weighted average and standard deviation respectively in the following manner:
\begin{equation}
\mu(\mathbf{Y},\mathbf{T})  = \frac{\sum_{i,j}{(\mathbf{Y} \odot \mathbf{T})_{i,j}}}{\epsilon + \sum_{i,j}{{\mathbf{T}_{i,j}}}}, 
\sigma(\mathbf{Y},\mathbf{T})  = \sqrt{\frac{\sum_{i,j}{(\mathbf{Y} \odot \mathbf{T} - \mu(\mathbf{Y},\mathbf{T}))^2_{i,j}}}{\epsilon + \sum_{i,j}{{\mathbf{T}_{i,j}}}}+\epsilon},
\end{equation}
where $\mathbf{Y}$ is a feature map, $\mathbf{T}$ is a soft mask with the same size of $\mathbf{Y}$, and $\epsilon$ is a small positive constant. $i$ and $j$ are the indexes of the height and width, respectively.

Literature \cite{gatys2016image,johnson2016perceptual} shows feature mean is related to its global semantic information and variance is highly correlated to local patterns like texture. The feature statistics propagation by PCN helps regenerate inconsistent areas by leveraging contextual mean and variance. 
This is intrinsically different from existing approaches that achieve this implicitly in deep layers, as we explicitly accomplish it in each building block.
Thus PCN is beneficial to the learning procedure and performance of blind inpainting. More importantly, RIN keeps robust considering potential errors brought by $\mathbf{\hat{M}}$ from MPN, although RIN is guided by $\mathbf{\hat{M}}$ for repairing. The supporting experiments are given in Section \ref{sec_ab}.

Other special design in RIN includes feature fusion and a comprehensive optimization target. Feature fusion denotes concatenating the discriminative feature (bottleneck of MPN) to the bottleneck of RIN. This not only enriches the given features to be transformed into a natural image by introducing potential spatial information, but also enhances the discriminative learning for the location problem based on the gradients from the generation procedure.

The learning objective of RIN considers pixel-wise reconstruction errors, the semantic and texture consistency, and a learnable optimization target by fooling a discriminator via generated images as
\begin{equation} 
\mathcal{L}_g(\mathbf{\hat{O}}, \mathbf{O}) =  \underbrace{\lambda_{r}||\mathbf{\hat{O}} - \mathbf{O}||_1}_{\text{reconstruction term}} + \underbrace{\lambda_{s}{||V_{\mathbf{\hat{O}}}^{l}-V_{\mathbf{O}}^{l}||_1}}_{\text{semantic consistency term}}  + \underbrace{\lambda_{f} \mathcal{L}_{mrf}(\mathbf{\hat{O}}, \mathbf{O})}_{\text{texture consistency term}} + \underbrace{\lambda_{a} \mathcal{L}_{adv}(\mathbf{\hat{O}}, \mathbf{O})}_{\text{adversarial term}},
\end{equation}
where $\mathbf{\hat{O}}= G(\mathbf{I} | \mathbf{\hat{M}})$. $V$ is a pre-trained classification network (VGG19). $V_{\mathbf{O}}^{l}$ means we extract the feature layer $l$ (ReLU3\_2) of the input $\mathbf{O}$ when $\mathbf{O}$ is passed into $V$. Besides, $\lambda_{r}$, $\lambda_{s}$, $\lambda_{f}$, and $\lambda_{a}$ are regularization coefficients to adjust each term influence, and they are set to 1.4, 1e-4, 1e-3, and 1e-3 in our experiments, respectively.

ID-MRF loss \cite{wang2018inpainting,mechrez2018contextual} is employed as our texture consistency term. It computes the sum of the patch-wise difference between neural patches from the generated content and those from the corresponding ground truth using a relative similarity measure. It enhances generated image details by minimizing discrepancy with its most similar patch from the ground truth.

For the adversarial term, WGAN-GP \cite{gulrajani2017improved,arjovsky2017wasserstein} is adopted as
\begin{equation} 
\mathcal{L}_{adv}(\mathbf{\hat{O}}, \mathbf{O})= -E_{\mathbf{\hat{O}} \sim \mathbb{P}_{\mathbf{\hat{O}}}}[D(\mathbf{\hat{O}})],
\end{equation}
where $\mathbb{P}$ denotes data distribution, and $D$ is a discriminator for the adversarial training. Its corresponding learning objective for the discriminator is formulated as
\begin{equation} 
\mathcal{L}_{D}(\mathbf{\hat{O}}, \mathbf{O})=E_{\mathbf{\hat{O}} \sim \mathbb{P}_{\mathbf{\hat{O}}}}[D(\mathbf{\hat{O}})]-E_{{\mathbf{O}} \sim \mathbb{P}_{\mathbf{O}}}[D(\mathbf{O})]+\lambda_{gp} E_{\tilde{{\mathbf{O}}}\sim \mathbb{P}_{\tilde{{\mathbf{O}}}}}[(||\nabla_{\tilde{{\mathbf{O}}}}D(\tilde{{\mathbf{O}}})||_2-1)^2],
\end{equation}
where $\tilde{{\mathbf{O}}}=t\mathbf{\hat{O}}+(1-t){\mathbf{O}}$, $t \in [0, 1]$, $\lambda_{gp}$ is set to 10 for stabilizing the adversarial training.

\subsection{Training Procedure} \label{sec_train}

Generation of training data is given in Eq. \eqref{degrade_model}, where the production of $\mathbf{M}$ is adopted from \cite{yu2019free} as free-form strokes. The final prediction of our model is $G(\mathbf{I} | F(\mathbf{I}))$. All input and output are linearly scaled within range $[-1,1]$.

There are two training stages. MPN and RIN are separately trained at first. After both networks are converged, we jointly optimize ${\min}_{\theta_{F}, \theta_{G}} \lambda_m \mathcal{L}_m (F(\mathbf{I}), \mathbf{M}) + \mathcal{L}_g (G(\mathbf{I} | F(\mathbf{I})), \mathbf{O})$ with $\lambda_m=2.0$.

\section{Experimental Results and Analysis}

Our model and baselines are implemented using Tensorflow (v1.10.1). The evaluation platform is a Linux server with an Intel Xeon E5 (2.60GHz) CPU and an NVidia TITAN X GPU. Our full model (MPN + RIN) has 3.79M parameters and costs around 41.64ms to process a $256 \times 256$ RGB image.

The datasets include FFHQ (faces) \cite{karras2018style}, CelebA-HQ (faces) \cite{karras2017progressive}, ImageNet (objects) \cite{deng2009imagenet}, and Places2 (scenes) \cite{zhou2017places}. Our training images are all with size $256 \times 256$ unless otherwise specified. For FFHQ, images are downsampled from the original $1024 \times 1024$. For ImageNet and Places2, central cropping and padding are applied. When training on FFHQ, its corresponding noisy images are drawn from the training sets of CelebA-HQ and ImageNet. For training on ImageNet and Places2, these two datasets are the noisy source for each other.

Our baselines are all based on GAN \cite{goodfellow2014generative} frameworks. We construct four alternative models to show the influence brought by the network architecture and module design. For a fair comparison, \emph{they are all equipped with mask prediction network (MPN) in front of their inputs, and trained from stratch} (their MPNs are trained in the same way explained in Section \ref{sec_train}). The first two are built upon the contextual attention (CA) model \cite{yu2018generative} and generative multi-column (GMC) model \cite{wang2018inpainting}.  The input of these two inpainting variants is the concatenation of the estimated soft mask and the noisy image. The last two baselines are by employing the partial convolution (PC) \cite{liu2018image} and gated convolution (GC) \cite{yu2019free} as their basic building block, respectively, to construct the network, intending to explore how the fundamental neural unit affects this blind inpainting. Compared with our VCN (3.79M), the model complexity of these baselines CA, GMC, PC, and GC are all higher as 4.86M, 13.7M, 4.69M, and 6.06M, respectively. All these numbers include the model complexity of MPN as 1.96M.

\subsection{Mask Estimation Evaluation} \label{sec_mask}

\begin{table*}[!t]
	\centering
	\caption{Quantitative results on the testing sets from different methods.}
	\label{tb_me_evaluation2}
	\small
	\begin{tabular}{c|ccccccccc}
		\hline
		\multirow{2}*{Method} & \multicolumn{3}{c}{FFHQ-2K}  & \multicolumn{3}{c}{Places2-4K} & \multicolumn{3}{c}{ImageNet-4K}\\
		~ & BCE$\downarrow$ & PSNR$\uparrow$ & SSIM$\uparrow$ & BCE$\downarrow$ & PSNR$\uparrow$ & SSIM$\uparrow$ & BCE$\uparrow$ & PSNR$\uparrow$ & SSIM$\uparrow$\\
		\hline
		CA \cite{yu2018generative} & $1.297$ & 16.56 & 0.5509 & $0.574$ & 18.12 & 0.6018 & $0.450$ & 17.68 & 0.5285\\
		GMC \cite{wang2018inpainting} & $0.766$ & 20.06 & 0.6675 & $0.312$ & 20.38 & 0.6956 & $0.312$ & 19.56 & $\mathbf{0.6467}$\\
		PC \cite{liu2018image} & $\mathbf{0.400}$ & 20.19 & 0.6795 & $0.273 $ & 19.73 & 0.6682 & $0.229$ & 19.53 & 0.6277\\
		GC \cite{yu2019free} & 0.660 & 17.16 & 0.5915 & 0.504 & 18.42 & 0.6423 & 0.410 & 18.35 & 0.6416\\
		Our VCN & $\mathbf{0.400}$ & $\mathbf{20.94}$ & $\mathbf{0.6999}$ & $\mathbf{0.253}$ & $\mathbf{20.54}$ & $\mathbf{0.6988}$ & $\mathbf{0.226}$ & $\mathbf{19.58}$ & 0.6339\\
		\hline
	\end{tabular}
\end{table*}

We evaluate the mask prediction performance of all used methods based on their computed binary cross entropy (BCE) loss (the lower the better) on the testing sets. Note mAP or PR-curve is not employed as in semantic segmentation or detection task, because these metrics are mainly for binary masks (while our generated ones are soft) under certain thresholds. 

As shown in Table \ref{tb_me_evaluation2}, our VCN achieves superior performance compared to GC \cite{yu2019free}, PC \cite{liu2018image}, GMC \cite{wang2018inpainting}, and CA \cite{yu2018generative}, except that our SSIM in ImageNet-4K is slightly lower than GMC. It shows that different generative structures and modules affect not only generation but also the relevant mask estimation performance. Clearly, VCN with spatial normalization works decently, benefiting mask prediction by propagating clean pixels to the damaged areas. 

Partial convolution (in PC \cite{liu2018image}) yields relatively lower performance, and direct concatenation between the estimated mask and the input image (used in CA \cite{yu2018generative} and GMC \cite{wang2018inpainting}) is least effective. Visual comparison of the predicted masks of different methods is included in Fig. \ref{fig_face_results}, where the results from PC and VCN are comparable. They look better than those of CA, GMC, and GC.

\subsection{Blind Inpainting Evaluation} \label{sec_exp}

\begin{figure}[t]
	\begin{center}
		\centering
		\begin{tabular}{cccccc}
			\multicolumn{6}{c}{\includegraphics[width=1\linewidth]{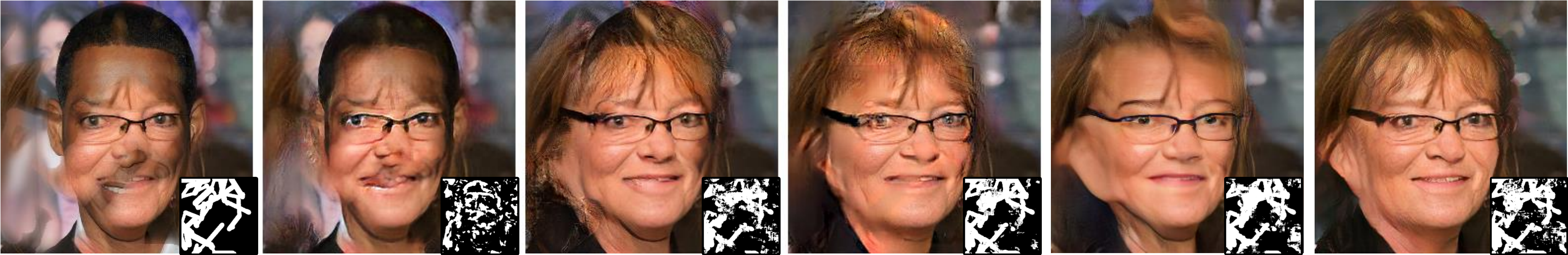}}\\
			\multicolumn{6}{c}{\includegraphics[width=1\linewidth]{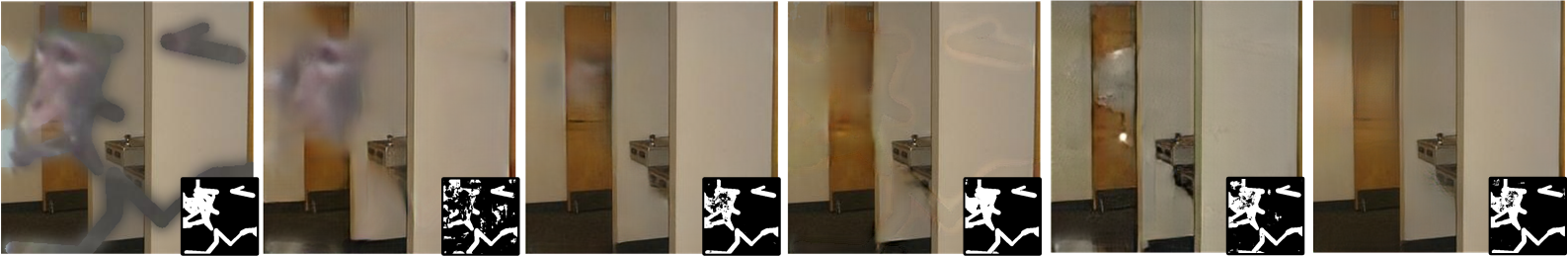}}\\
			\multicolumn{6}{c}{\includegraphics[width=1\linewidth]{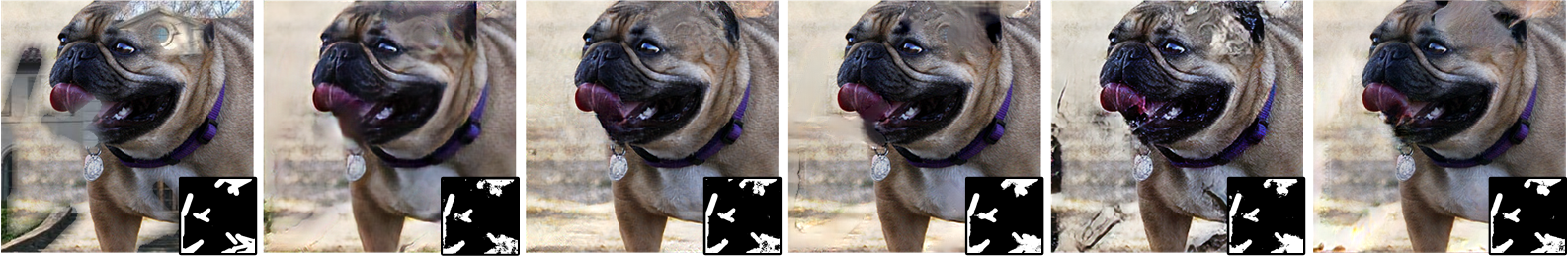}}\\
			\hspace{0.08\columnwidth}(a) & \hspace{0.12\columnwidth}(b) & \hspace{0.14\columnwidth}(c) & \hspace{0.14\columnwidth}(d) & \hspace{0.12\columnwidth}(e) & \hspace{0.08\columnwidth}(f)\\
		\end{tabular}
	\end{center}
		\vspace{-0.15in}
	\caption{Visual comparison on synthetic data (random stroke masks) from FFHQ (top), Places2 (middle), and ImageNet (bottom). (a) Input image. (b) CA \cite{yu2018generative}. (c) GMC \cite{wang2018inpainting}. (d) PC \cite{liu2018image}. (e) GC \cite{yu2019free}. (f) Our results. The ground truth masks (shown in the first column) and the estimated ones (in binary form) are shown on the bottom right corner of each image. More comparison is given in the supplementary file.}
	\label{fig_face_results}
\end{figure}
	
\begin{figure}[t]
	\begin{center}
		\centering
		\begin{tabular}{cccc}
			\multicolumn{2}{c}{\includegraphics[width=0.48\linewidth]{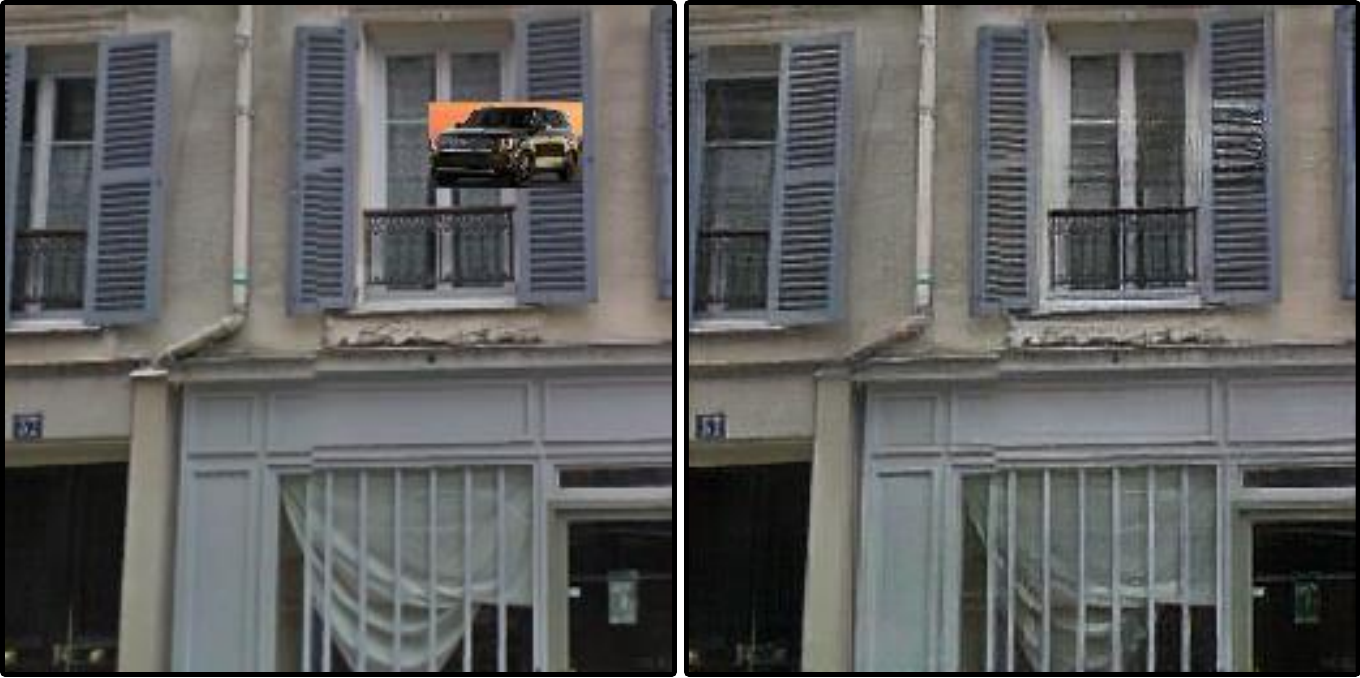}} & \multicolumn{2}{c}{\includegraphics[width=0.48\linewidth]{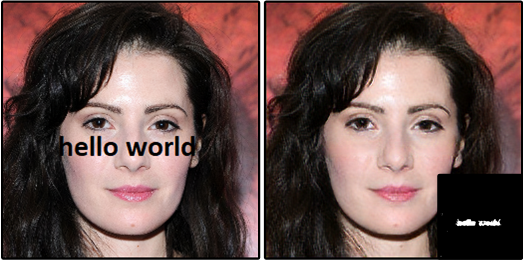}}\\
			
			\hspace{0.1\columnwidth}(a) & \hspace{0.08\columnwidth}(b) & \hspace{0.1\columnwidth}(a) & \hspace{0.08\columnwidth}(b) \\
		\end{tabular}
	\end{center}
	\vspace{-0.15in}
	\caption{Results from VCN on facades and faces with other shaped masks: (a) input, and (b) results.}
	\label{fig_syn_result2}
\end{figure}

\begin{figure*}[t]
	\begin{center}
		\centering
		\begin{tabular}{cccccccc}
			\multicolumn{8}{c}{\includegraphics[width=1\linewidth]{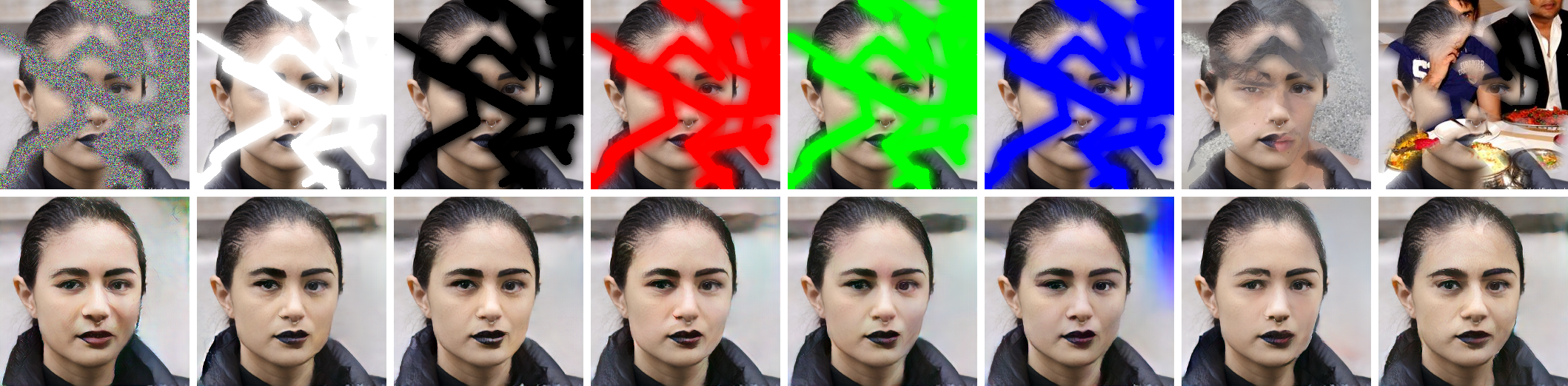}}\\
		\end{tabular}
	\end{center}
	\vspace{-0.15in}
	\caption{Visual evaluations on FFHQ with random masks filled with different content. First row: input; second row: corresponding results from our model. Last two images are filled with content drawn from the testing sets of CelebA-HQ and ImageNet respectively.}
	\label{fig_face_robust}
\end{figure*}

\vspace{-0.1in}
\paragraph{\textbf{Synthetic Experiments}}
Visual comparisons of the used baselines and our method on the synthetic data (composed in the way we give in Sec. \ref{sec_data_gen}) are given in Fig. \ref{fig_face_results}. Our method produces more visually convincing results with fewer artifacts, which are not much disturbed by the unknown contamination areas. On the other hand, the noisy areas from CA and GMC baselines manifest that concatenation of mask and input to learn the ground truth is not an effective way for blind inpainting setting. More examples are given in the supplementary file.

\begin{figure}[!ht]
	\begin{center}
		\centering
		\begin{tabular}{cccc}
			\multicolumn{2}{c}{\includegraphics[width=0.48\linewidth]{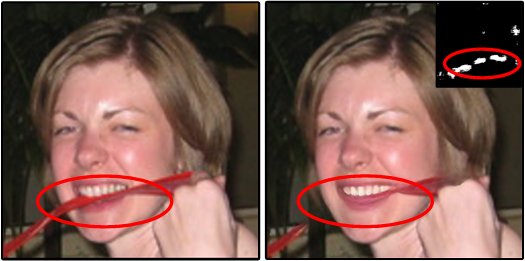}} & \multicolumn{2}{c}{\includegraphics[width=0.48\linewidth]{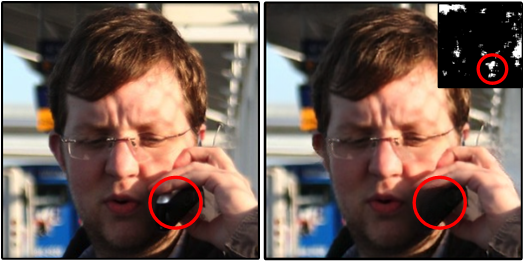}}\\
			\hspace{0.1\columnwidth}(a) & \hspace{0.06\columnwidth}(b) & \hspace{0.1\columnwidth}(a) & \hspace{0.06\columnwidth}(b) \\
		\end{tabular}
	\end{center}
	\vspace{-0.15in}
	\caption{Blind inpainting on the real occluded faces from COCO dataset with VCN: (a) input, and (b) results. Red ellipses in the pictures highlight the regions to be edited.}
	\label{fig_real_result2}
\end{figure}

About randomly inserted patches or text shape masks, Fig. \ref{fig_syn_result2} shows that our method can locate the inserted car, complete the facade (train/test on Paris Streetview \cite{pathak2016context}), and restore text-shape corrupted regions on the testing face image from FFHQ.

\vspace{-0.1in}
\paragraph{\textbf{Robustness against Various Degradation Patterns}} Our training scheme makes the proposed model robust to filling content, as shown in Fig. \ref{fig_face_robust}. It can deal with Gaussian noise or constant color filling directly, while these patterns are not included in our training. This also shows such a training scheme makes the model learn to tell and inpaint the visual inconsistency instead of memorizing the synthetic missing data distribution.

\begin{table}
	\centering
	\caption{User studies. Each entry gives the percentage of cases where results by our approach are judged as more realistic than another solution. The observation and decision time for users is unlimited.}
	\label{tb_user_studies}
	\begin{tabular}{c|cccc}
		\hline
		Methods & VCN $>$ CA & VCN $>$ GMC & VCN $>$ PC & VCN $>$ GC\\
		\hline
		FFHQ & 99.64\% & 80.83\% & 77.66\% &  92.15\%\\
		Places2  & 81.39\% & 51.63\% & 70.49\% & 78.15\%\\
		ImageNet& 91.20\% & 50.09\% & 77.92\% & 83.30\%\\
		\hline
	\end{tabular}
	\vspace{-0.1in}
\end{table}

PSNR and SSIM indexes evaluated on the testing sets of the used datasets are given in Table \ref{tb_me_evaluation2} for reference. Generally, VCN yields better or comparable results compared with baselines, verifying the effectiveness of spatial normalization about image fidelity restoration in this setting.

Further, pairwise A/B tests are adopted for blind user studies using Google Forms. 50 participants are invited for evaluating 12 questionnaires. Each has 40 pairwise comparisons, and every comparison shows the results predicted from two different methods based on the same input, randomized in the left-right order. As given in Table \ref{tb_user_studies}, our method outperforms the CA, PC, and GC in all used datasets and GMC in FFHQ, and yields comparable visual performance with GMC on ImageNet and Places2 with a much smaller model size (3.79M vs. 13.70M).

\begin{figure}[!t]
	\begin{center}
		\centering
		\small
		\begin{tabular}{ccc}
			\multicolumn{3}{c}{\includegraphics[width=0.95\linewidth]{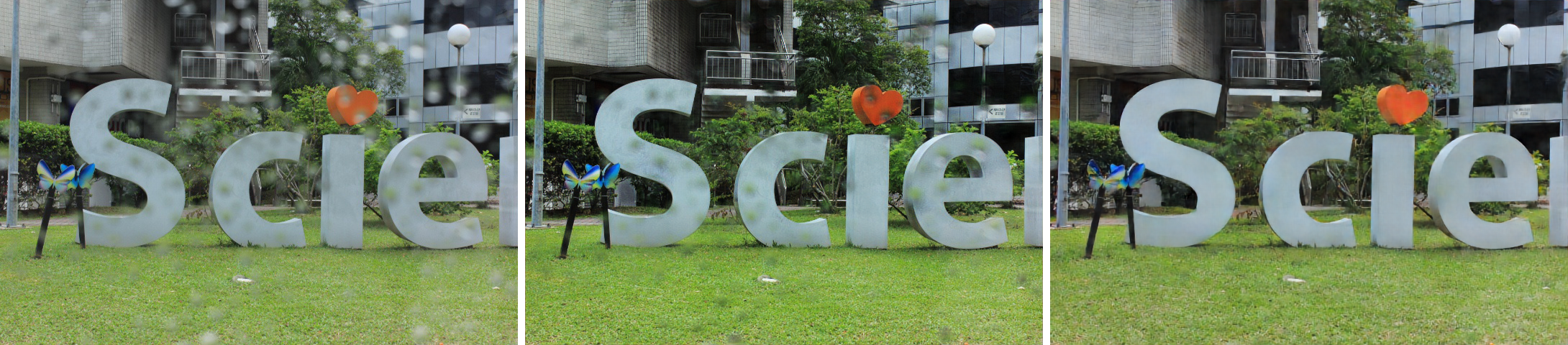}}\\
			\hspace{0.15\columnwidth}(a) & \hspace{0.28\columnwidth}(b) &
			\hspace{0.14\columnwidth}(c)\\
		\end{tabular}
	\end{center}
	\vspace{-0.15in}
	\caption{Visual evaluation on raindrop removal dataset. (a) Input image. (b) AttentiveGAN \cite{qian2018attentive}. (c) Ours (Best view in original resolution). More examples are in the supplementary file.}
	\label{fig_raindrop_results}
\end{figure}


\vspace{-0.1in}
\paragraph{\textbf{Blind Inpainting on Real Cases}}  Fig. \ref{fig_real_result2} gives blind inpainting (trained on FFHQ) on the occluded face from COCO dataset \cite{lin2014microsoft}. Note VCN can automatically, and at least partially, restore these detected occlusions. The incomplete removal with red strip bit in the mouth may be caused by similar patterns in FFHQ, as mentioned that the detected visual inconsistency is inferred upon the learned distribution from the training data. 

\vspace{-0.1in}
\paragraph{\textbf{Model Generalization}} We evaluate the generality of our model on raindrop removal with a few training data. The dataset in \cite{qian2018attentive} gives paired data (noisy and clean ones) without masks. Our full model (pre-trained on Places2 with random strokes) achieves promising qualitative results (Fig. \ref{fig_raindrop_results}) on the testing set, which is trained with a few training images (20 RGB images of resolution $480 \times 720$, around 2.5\% training data). In the same training setting, testing results by AttentiveGAN \cite{qian2018attentive} (a raindrop removal method) yield 24.99dB while ours is 26.22dB. It proves the learned visual consistency ability can be transferred to other similar removal tasks with a few target data.

\subsection{Ablation Studies} \label{sec_ab}

\begin{table}[!t]
	\centering
	\caption{Quantitative results of component ablation of VCN on FFHQ dataset (ED: Encoder-decoder; fusion: the bottleneck connection between MPN and RIN; -RM: removing the estimated contamination as $G(\mathbf{I} \odot (\mathbf{1}-\hat{\mathbf{M}})|\hat{\mathbf{M}})$; SC: semantic consistency term).}
	\label{tb_index_ab}
	\begin{tabular}{c|cccccc}
		\hline
		Model & ED & VCN w/o MPN & VCN w/o fusion & VCN w/o SC & VCN-RM & VCN full\\
		\hline
		PSNR$\uparrow$ & 19.43 & 18.88 & 20.06 & 20.56 & 20.87 & \textbf{20.94}\\
		SSIM$\uparrow$ & 0.6135 & 0.6222 & 0.6605 & 0.6836 & \textbf{0.7045} & 0.6999 \\
		BCE$\downarrow$ & - & - & 0.560 & 0.653 & 0.462 & \textbf{0.400}\\
		\hline
	\end{tabular}
\end{table}

\begin{figure}
	\begin{center}
		\centering
		\begin{tabular}{cccccc}
			\multicolumn{6}{c}{\includegraphics[width=1\linewidth]{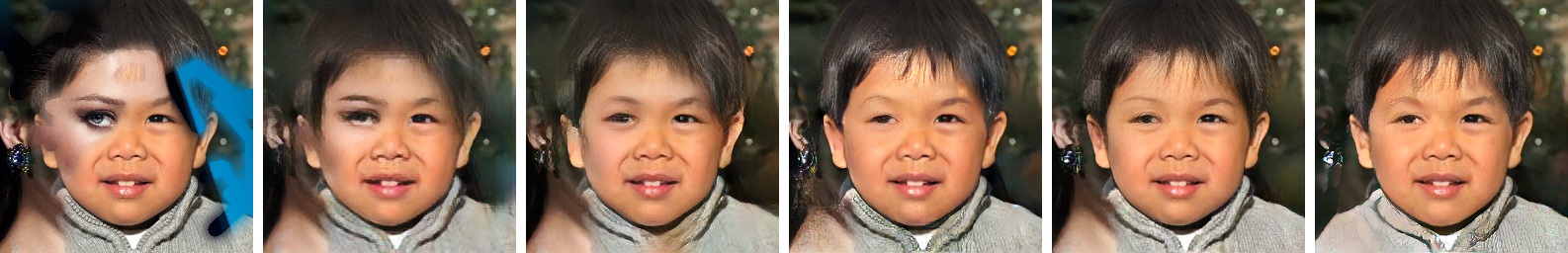}}\\
			\hspace{0.07\columnwidth}(a) & \hspace{0.13\columnwidth}(b) & \hspace{0.13\columnwidth}(c) & \hspace{0.13\columnwidth}(d) & \hspace{0.13\columnwidth}(e) & \hspace{0.08\columnwidth}(f)\\
		\end{tabular}
	\end{center}
	\vspace{-0.15in}
	\caption{Visual comparison on FFHQ using VCN variants. (a) Input image. (b) VCN w/o MPN. (c) VCN w/o skip. (d) VCN w/o semantics. (e) VCN-RM. (f) VCN full.}
	\label{fig_ablation}
\end{figure}

\begin{figure} [!ht]
	\begin{center}
		\centering
		\begin{tabular}{cccccc}
			\multicolumn{6}{c}{\includegraphics[width=1\linewidth]{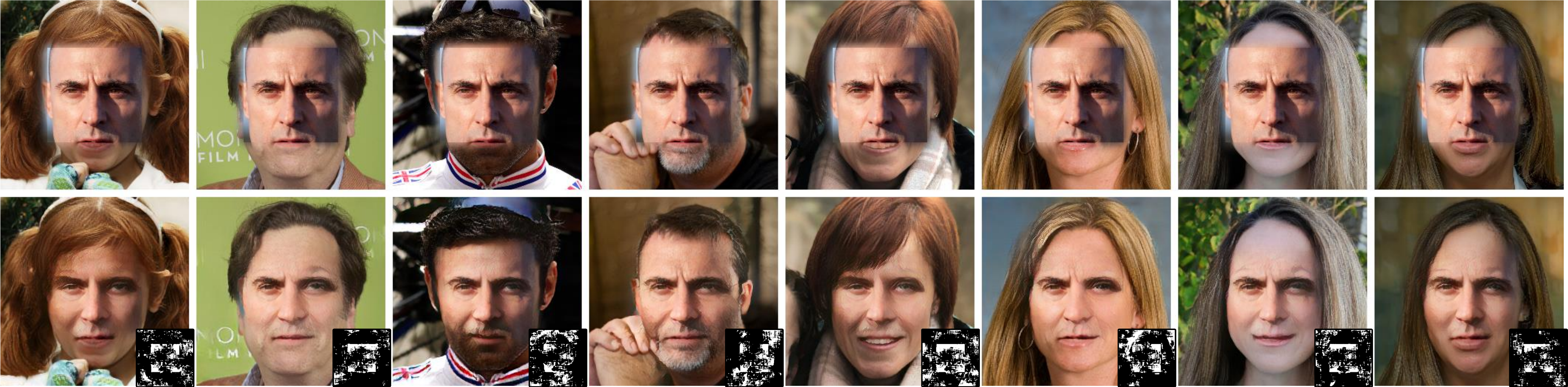}}\\
		\end{tabular}
	\end{center}
	\vspace{-0.15in}
	\caption{Visual editing (face-swap) on FFHQ. First row: image with coarse editing where a new face (from CelebA-HQ) is pasted at the image center; Second row:  corresponding results from our model. Best viewed with zoom-in.}
	\label{fig_face_swap}
\end{figure}


\vspace{-0.1in}
\paragraph{\textbf{W and W/O MPN}} Without MPN, the fidelity restoration of VCN will degrade a lot in Table \ref{tb_index_ab}. The comparison in Fig. \ref{fig_ablation}(b) shows VCN w/o MPN can find some obvious artifacts like blue regions, but it fails to completely remove the external introduced woman face. Thus our introduced task decomposition and joint training strategy are effective. Compared with the performance of ED, VCN variants show the superiority of the module design in our solution.

\vspace{-0.1in}
\paragraph{\textbf{Fusion of Discriminative and Generative Bottlenecks}} The improvement of such modification on mask prediction (BCE), PSNR, and SSIM is limited. But this visual improvement shown in Fig. \ref{fig_ablation}(c) and (f) is notable. Such shortcut significantly enhances detail generation.

\vspace{-0.1in}
\paragraph{\textbf{Input for Inpainting}} Since the filling mask is estimated instead of being given, removing the possible contamination areas may degrade the generation performance due to the error accumulation caused by mask prediction. Fig. \ref{fig_ablation}(e) validates our consideration.

\vspace{-0.1in}
\paragraph{\textbf{Loss Discussion}} The significance of the semantic consistency term affects VCN is given in Table \ref{fig_ablation}. It shows this term benefits the discrimination ability and fidelity restoration of the model since removing it leads to a decrease of PSNR (from 20.94 to 20.56) and SSIM (from 0.6999 to 0.6836), and increase of BCE (from 0.4 to 0.653). Removing this term would lead to hair and texture artifacts like that in Fig. \ref{fig_ablation}(d). Other terms have been discussed in \cite{yu2018generative,wang2018inpainting}. 

\vspace{-0.1in}
\paragraph{\textbf{Study of PCN}} In the testing phase, we adjust $\rho$ (instead of using the trained one) manually in PCN to show its flexibility in controlling the interference of possible contamination (in the supplementary file). With the increase of $\rho$, VCN tends to generate missing parts based on context instead of blending the introduced `noise'.

\begin{figure}[!t] 
	\begin{center}
		\centering
		\begin{tabular}{cccccc}
			\multicolumn{6}{c}{\includegraphics[width=0.98\linewidth]{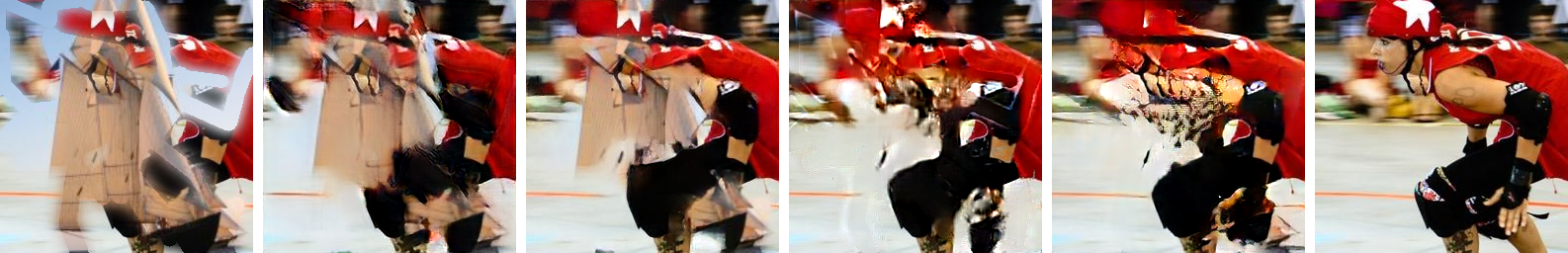}}\\
			\hspace{0.07\columnwidth}(a) & \hspace{0.13\columnwidth}(b) & \hspace{0.13\columnwidth}(c) & \hspace{0.13\columnwidth}(d) & \hspace{0.13\columnwidth}(e) & \hspace{0.08\columnwidth}(f)\\
		\end{tabular}
	\end{center}
	\vspace{-0.15in}
	\caption{Failure cases on Places2. (a) Input image. (b) CA \cite{yu2018generative}. (c) GMC \cite{wang2018inpainting}. (d) PC \cite{liu2018image}. (e) Our results. (f) Ground truth.}
	\label{fig_failure_case} 
\end{figure}

\vspace{-0.1in}
\paragraph{\textbf{Applications on Image Blending}} \label{sec_face_swap}
Our blind inpainting system also finds applications on image editing, especially on blending user-fed visual material with the given image automatically. We attribute this application to the adversarial term in our formulation, which just encourages the generated result to be realistic as the natural ones. From this perspective, our method can utilize the filling content to edit the original ones. The given materials from external datasets will be adjusted on its shape, color, shadow, and even its semantics to appeal to its new context, as given in Fig. \ref{fig_face_swap}. The editing results are natural and intriguing. Note the estimated masks mainly highlight the outlines of the pasted rectangle areas, which are just inconsistent regions according to context. 

\vspace{-0.1in}
\paragraph{\textbf{Limitation and Failure Cases}}
If contaminated areas in images are large enough to compromise main semantics, our model cannot decide which part is dominant and the performance would degrade dramatically. As shown in Fig. \ref{fig_failure_case}, our model cannot produce reasonable results when the majority of the input images are damaged. Moreover, if users want to remove a certain object from an image, it would be better to feed the user’s mask into the robust inpainting network to complete the target regions. On the other hand, our method cannot repair the common occlusion problems (like human body occlusion) because our model does not regard this as an inconsistency. 

\section{Conclusion}

We have proposed a robust blind inpainting framework with promising restoration ability on several benchmark datasets. We designed a new way of data preparation, which relaxes missing data assumptions, as well as an end-to-end model for joint mask prediction and inpainting. A novel probabilistic context normalization is used for better context learning. 
Our model can detect incompatible visual signals and transform them into contextual consistent ones. It is suitable to automatically repair images when manually labeling masks is hard. Our future work will be to explore the transition between traditional inpainting and blind inpainting, \eg\ using coarse masks or weakly supervised hints to guide the process.

%
%
\bibliographystyle{splncs04}
\bibliography{egbib}
\end{document}